\documentclass[letterpaper, 10 pt, conference]{ieeeconf}  %

\IEEEoverridecommandlockouts                              %

\makeatletter
\let\NAT@parse\undefined
\makeatother

\usepackage{graphics} %
\usepackage{graphicx}
\usepackage{grffile} %
\usepackage{amsmath} %
\usepackage{amssymb}  %
\usepackage{color}
\usepackage{bm}
\usepackage{url}

\usepackage[noend]{algpseudocode}
\usepackage{algorithm}
\usepackage{makecell}
\usepackage{subfigure}
\usepackage[square, comma, numbers,sort&compress]{natbib}
\usepackage{multirow}
\let\labelindent\relax
\usepackage{enumitem}
\usepackage{pifont}
\newcommand{\xmark}{\ding{55}}%
\usepackage[table,dvipsnames]{xcolor}
\usepackage[pagebackref=true,breaklinks=true,colorlinks,bookmarks=false]{hyperref}

\hypersetup{
linkcolor=BrickRed
,citecolor=Green
,filecolor=Mulberry
,urlcolor=NavyBlue
,menucolor=BrickRed
,runcolor=Mulberry
,linkbordercolor=BrickRed
,citebordercolor=Green
,filebordercolor=Mulberry
,urlbordercolor=NavyBlue
,menubordercolor=BrickRed
,runbordercolor=Mulberry
}

\def\ourname{NYU-VPR}

\title{\LARGE \bf
\ourname: Long-Term Visual Place Recognition Benchmark\\with View Direction and Data Anonymization Influences
}

\author{\small Diwei Sheng$^{*}$, Yuxiang Chai$^{*}$, Xinru Li, Chen Feng$^{\dagger}$, Jianzhe Lin, Claudio Silva, John-Ross Rizzo\\% <-this %
\url{https://ai4ce.github.io/NYU-VPR/}
\thanks{New York University, Brooklyn, NY 11201, USA}%
\thanks{$^{*}$ indicates equal contributions.}%
\thanks{$^{\dagger}$Chen Feng is the corresponding author.
    {\tt\small cfeng@nyu.edu}}%
}

\begin{document}

\maketitle
\thispagestyle{empty}
\pagestyle{empty}

\begin{abstract}

Visual place recognition (VPR) is critical in not only localization and mapping for autonomous driving vehicles, but also assistive navigation for the visually impaired population. To enable a long-term VPR system on a large scale, several challenges need to be addressed. First, different applications could require different image view directions, such as front views for self-driving cars while side views for the low vision people. Second, VPR in metropolitan  scenes can often cause privacy concerns due to the imaging of pedestrian and vehicle identity information, calling for the need for data anonymization before VPR queries and database construction. Both factors could lead to VPR performance variations that are not well understood yet. To study their influences, we present the \ourname~dataset that contains more than 200,000 images over a 2km$\times$2km area near the New York University campus, taken within the whole year of 2016. We present benchmark results on several popular VPR algorithms showing that side views are significantly more challenging for current VPR methods while the influence of data anonymization is almost negligible, together with our hypothetical explanations and in-depth analysis.
\footnote{\textcolor{blue}{After IROS'21, Manuel Lopez Antequera (mlop@fb.com) points out that our description of Mapillary Street-Level Sequences dataset (MSLS) is inaccurate. We modify this paper accordingly (highlighted in blue), although the main conclusions are not changed.}}

\end{abstract}

\section{Introduction}

\begin{table*}[t]
    \caption{Comparison of major public outdoor VPR datasets with \ourname.}
    \vspace{-3mm}
    \label{tab_1}
    \centering
    \resizebox{\textwidth}{!}{%
    \begin{tabular}{l|ccccccc}
    \hline
    Dataset     & side-view & side-view-label & dynamic-object & crowded-area & anonymization & seasonal-changes & \#images\\ \hline
    
    StreetLearn~\cite{mirowski2019streetlearn}& \checkmark & - & \checkmark & \checkmark & \xmark & \xmark & 143,000\\\hline
    
    StreetView~\cite{zamir2014image}& \checkmark & - & \checkmark & \checkmark & \xmark & \xmark & 62,058\\\hline
    
    Nordland~\cite{sunderhauf2013we}& \xmark & - & \xmark & \xmark & \xmark & \checkmark & 28,865\\ \hline  
    
    VPRiCE 2015~\cite{Suenderhauf_vprice_2015} & \xmark & -  & \checkmark & \xmark & \xmark & \xmark & 7,778\\ \hline
    
    Tokyo 24/7~\cite{torii201524} & \checkmark & \xmark & \checkmark & \checkmark & face-only & \xmark &76,000\\ \hline
    
    Pittsburgh~\cite{Torii-CVPR2013} & \checkmark & \xmark & \checkmark & \xmark & \xmark & \checkmark  & 254,064\\ \hline
     
    KITTI raw~\cite{Geiger2013IJRR}& \xmark & - & \checkmark & \xmark & \xmark & \xmark &12,919\\ \hline
    
    KAIST~\cite{parkall} & \xmark & - & \checkmark & \xmark & \xmark & \xmark &105,000\\ \hline
    
    Oxford RobotCar~\cite{maddern20171} & \xmark & - & \checkmark & \xmark & \xmark & \checkmark & 19,556,490\\ \hline
    
    \textcolor{blue}{MSLS}~\cite{warburg2020mapillary} & \checkmark & \checkmark & \checkmark & \checkmark & \textcolor{blue}{\checkmark(blurry)} & \checkmark & 1,681,000\\ \hline
    
    NCLT~\cite{carlevaris2016university} & \xmark & - & \checkmark & \xmark & \xmark & \checkmark & 100,000\\ \hline
    
    \textbf{\ourname (ours)} & \checkmark & \checkmark & \checkmark & \checkmark & \textcolor{blue}{\checkmark(erase)} & \checkmark & 201,790\\ \hline

    \end{tabular}%
    }
    \vspace{-5mm}
\end{table*}

Visual place recognition (VPR) is the process of retrieving the most similar images for a query one from a database of images with known camera poses, which is often used for loop closing in mapping, localization, and navigation. It relies on representing an image as a global feature vector which describes the portion of the image appearance that is most relevant to its capturing pose. Its applications range from autonomous driving for vehicles, to assistive navigation for the visually impaired people, especially in busy and crowded metropolitan areas where GPS could suffer from the ``urban canyon'' problem when satellite signals are blocked or multi-reflected to cause large localization errors.

\begin{figure}[t]
    \centering
    \includegraphics[width=\columnwidth]{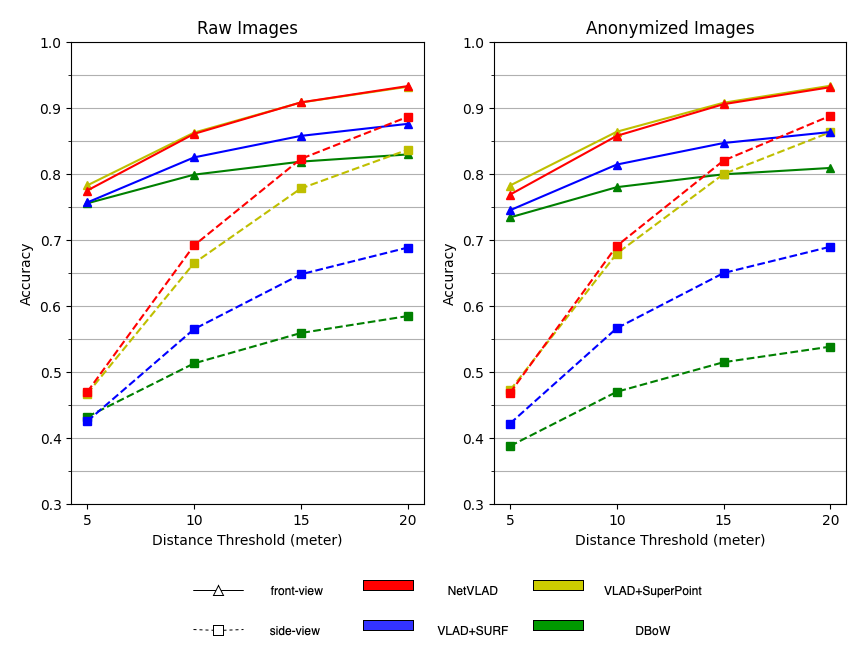}
    \vspace{-5mm}
    \caption{Top-1 VPR retrieval accuracy of 4 baseline algorithms under different view directions and data anonymization.}
    \label{fig:main}
    \vspace{-6mm}
\end{figure}

A reliable large-scale and long-term VPR system has to address several challenges. The first one is to choose a proper image view direction, \textit{assuming a non-omnidirectional camera which is commonly available on smartphones unlike panoramic cameras}. In a self-driving car scenario, using front-view images from a dash-mounted camera whose view direction is parallel to the driving/street direction is almost a default choice. In fact, most VPR methods have been investigated and evaluated under such front-view conditions, where features of roads, shapes of skylines, and textures of the roadside buildings can all contribute to describing and discriminating various image locations.

Yet not all downstream applications prefer front-view images. For example, a person with low vision might need image-based wearable navigation assistance to find the entrance of a particular shop on the street. The aforementioned front-view images typically contain more than half of the pixels on the road and the sky while the remaining pixels on street sides that are far away from the image capturing locations. Contrarily, the side view offers fronto-parallel images of buildings along streets, which stores enough information required for such an application scenario.

However, currently, there is a lack of datasets that could evaluate VPR methods specifically on side-view images in comparison with front-view ones. Many datasets contain no side-view images or mix them with front-view images without explicit labels (see Table~\ref{tab_1}). Moreover, as far as we know, there is no systematic comparison of VPR performance between images from the two view directions. Thus, the following questions remain unclear:
\textit{are side-view images more challenging for VPR than front-view ones? And if so, why? And how much is the performance difference?}

Another challenge for large-scale VPR systems is data privacy, receiving an increasing attention from the community~\cite{speciale2019privacy}. Building such systems in large metropolitan areas requires collecting images for a long term, inevitably creating concerns of both violating the privacy of the identity information of individual pedestrians and vehicles, and potentially even leaking their spatial-temporal trajectories. Unlike the privacy-preserving technique in~\cite{speciale2019privacy} that still operates on raw images, another way could be directly anonymizing the images by wiping out all the identity-related pixels (see Figure~\ref{fig:raw_anony}). However, this again brings some unanswered questions for VPR: \textit{would such data anonymization significantly affects the performance of existing VPR algorithms? If so, does it increase or decrease the VPR accuracy/robustness?}

This paper aims at filling the gaps by a new VPR dataset and benchmark. This is a year-long dataset captured outdoors by vehicles with front-view and side-view cameras traveling in a metropolitan area recording more than 200,000 GPS-tagged images. This allows us to answer the above questions by comparing the results under various conditions.

The contributions of this paper are as follows:
\begin{itemize}
\item We present \ourname, \textit{a unique large-scale, year-long, outdoor VPR benchmark dataset containing both front-view and side-view GPS-tagged images taken at different lighting conditions with seasonal and appearance changes in a busy and crowded urban area} of New York City. This dataset and our benchmark code will be released for educational and research purposes.
\item We benchmark the performance of several popular VPR algorithms with a focus on the influence of image view directions. As far as we know, this is \textit{the first work to systematically demonstrate and analyze the causes of the significant challenge of VPR with side-view images}.
\item We anonymize the identity information in this dataset by removing pixels of both pedestrians and vehicles to address the privacy concerns of large-scale VPR in urban scenes. This is also the first result to show that all the benchmarked VPR algorithms are only marginally affected by this anonymization.

\end{itemize}

\section{Related Work}

Because \ourname\ contains only outdoor images used for visual place recognition, we review publicly available datasets that have similar characteristics. The main differences between those datasets and our proposed dataset are summarized in Table \ref{tab_1}.

\textbf{Side-view and side-view label:} In recent years there has been substantial growth in the number of visual place recognition datasets in the urban areas. However, most datasets only contain front-view images, gathered by cameras on the front and back of cars~\cite{sunderhauf2013we,Suenderhauf_vprice_2015,Geiger2013IJRR,parkall,maddern20171,carlevaris2016university}. The side-view images featuring the storefront and sidewalk are not included. Few datasets include the side-view images in addition to the front-view images. For example, the images in Tokyo 24/7 dataset were gathered by pedestrians' phones and featured both front-view and side-view images~\cite{torii201524}. The images in Pittsburgh dataset were perspective images generated from Google Street View panoramas of the Pittsburgh area~\cite{Torii-CVPR2013}. But those datasets do not label the images as front-view or side-view. Thus no work can use those datasets to compare the visual place recognition results on the side-view images versus those on the front-view images. \ourname\ contains images labeled as front-view and images labeled as side-view. We focus on evaluating the visual place recognition algorithms in both categories. We compare the results of algorithms on the side-view images versus the results on the front-view images in order to analyze the influence of view direction on the long-term visual place recognition.

\textbf{Dynamic objects, crowded-area and anonymization:} Dynamic objects such as pedestrians and vehicles in images may affect the performance of visual place recognition due to the changing appearance at the same place or the existence of obstructing the street view. Besides, the presence of pedestrians and vehicles in publicly available datasets may raise privacy issues if images are not anonymized. There are few appearances of dynamic objects in datasets of images gathered in suburban areas. For example, Nordland is a dataset of images taken on the train on a railway line between the cities of Trondheim and Bodø~\cite{sunderhauf2013we}. In contrast, dynamic objects appears much more frequently in the datasets featuring urban areas~\cite{Suenderhauf_vprice_2015,torii201524,Torii-CVPR2013,Geiger2013IJRR,parkall,maddern20171,warburg2020mapillary}. We define crowded areas as metropolitan areas such as New York City and Tokyo that has a high population density and is crowded with pedestrians and vehicles. In Table \ref{tab_1}, images from Tokyo 24/7, \textcolor{blue}{MSLS}, and \ourname\ are gathered in crowded areas. Anonymization is needed on those datasets for privacy protection. \textcolor{blue}{Tokyo 24/7~\cite{torii201524} only applied face redaction on pedestrians, and MSLS~\cite{warburg2020mapillary} only anonymize faces and license plates.} We use MSeg~\cite{MSeg_2020_CVPR} to replace pedestrians and vehicles with white pixels.

\begin{figure}[t]
    \centering
    \subfigure[4 locations (row) at 4 seasons.
    ]{
        \includegraphics[width=0.48\linewidth]{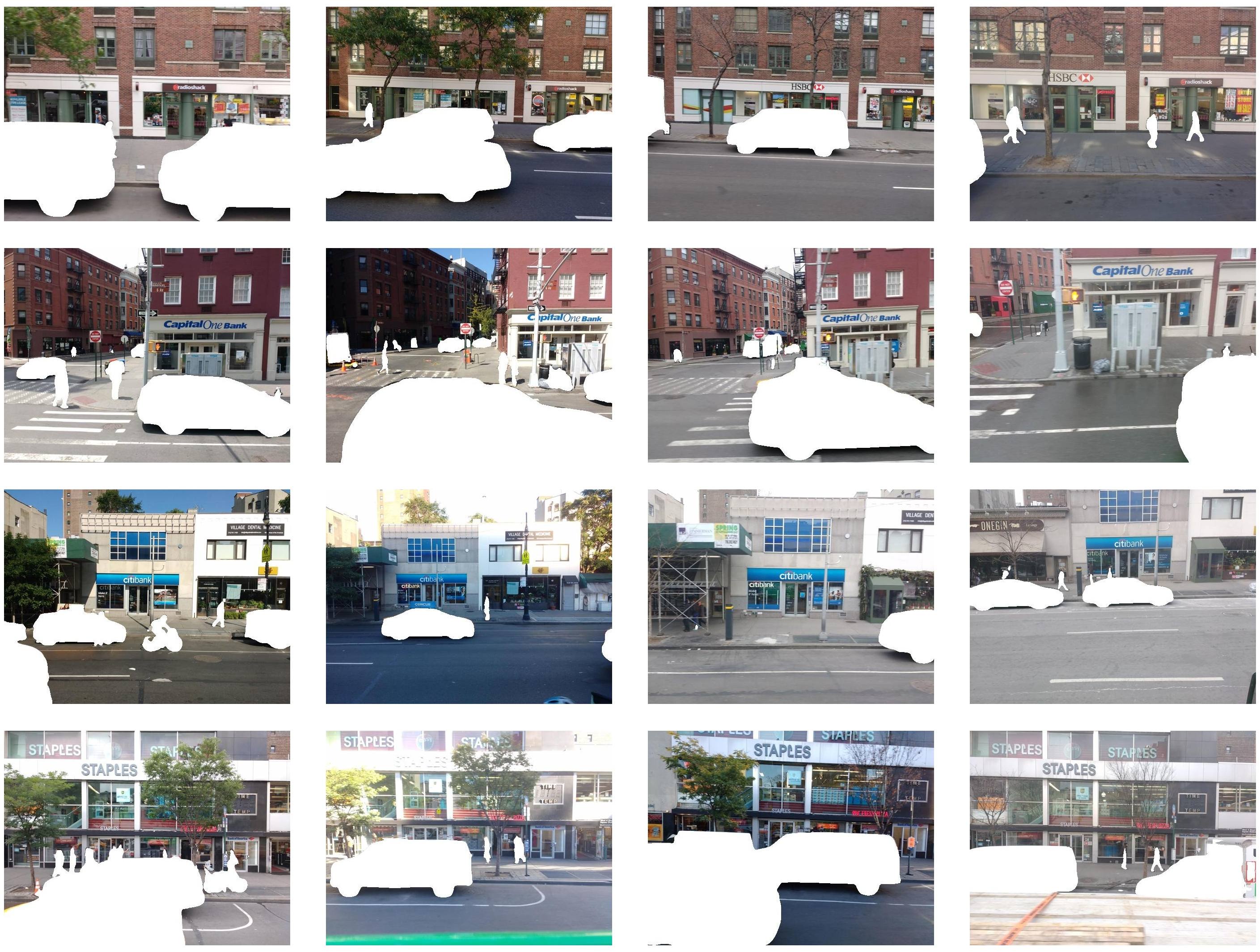}
        \label{fig:same_location}
    }
    \subfigure[Month/hour distributions.]{
        \includegraphics[width=0.45\linewidth]{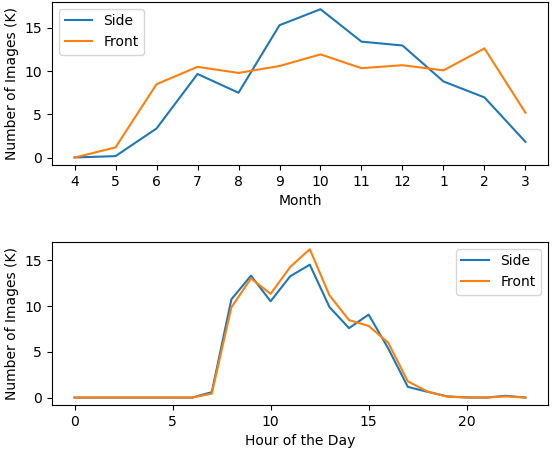}
        \label{fig:time_figure}
    }
    \caption{Dataset visualization of \ourname~w.r.t. the image capturing time.}
    \vspace{-6mm}
\end{figure}

\textbf{Seasonal changes:} Matching images that are taken at the same location in different seasons is crucial for long-term visual place recognition. This is because objects on images change with the seasons: new storefront, trees withering, constructions finished, etc. In Table \ref{tab_1}, Pittsburgh~\cite{Torii-CVPR2013} includes images in different seasons but few image locations are visited in all four seasons. Nordland~\cite{sunderhauf2013we}, Oxford RobotCar~\cite{maddern20171}, NCLT~\cite{carlevaris2016university}, and \textcolor{blue}{MSLS}~\cite{warburg2020mapillary} all include images in four seasons for most locations. \ourname\ is similar but temporally denser: most locations are visited more than once every month, as shown by the time (Fig.~\ref{fig:time_figure}) and space (Fig.~\ref{fig:Frequency_of_side_view} and Fig.~\ref{fig:Frequency_of_front_view}) distributions of images. So \ourname\ can be used to evaluate long-term visual place recognition for the influence of seasonal changes.

\textbf{Baseline methods:} Current VPR methods can be roughly grouped into three categories: deep-learning methods, non-deep-learning methods, and methods that only use deep learning descriptors. We select methods in each category. Deep learning methods~\cite{arandjelovic2016netvlad,kendall2015posenet,chancan2020hybrid,chen2017deep} use a convolutional neural networks (CNN) and train CNN in an end-to-end manner directly. We select NetVLAD~\cite{arandjelovic2016netvlad} and PoseNet~\cite{kendall2015posenet} in this group. For non-deep-learning methods~\cite{GalvezTRO12,jegou2010aggregating,sattler2012improving,torii201524}, two classical ones are bag-of-words (BoW) model and Vector of Locally Aggregated Descriptors (VLAD). In this group, we choose DBoW+ORB~\cite{GalvezTRO12,orb2011}, which was used in the popular ORB-SLAM for loop closing~\cite{7219438}, and VLAD+SURF~\cite{jegou2010aggregating, bay2006surf}, which was used in~\cite{yu2018vlase}. Methods that only use deep learning descriptors take advantage of deep nets' ability to detect a richer set of key points, such as SuperPoint~\cite{detone18superpoint} which we also adopted.

\section{The \ourname~Dataset}

\begin{figure}[htp]
\vspace{-2mm}
    \centering
    \includegraphics[width=8.5cm]{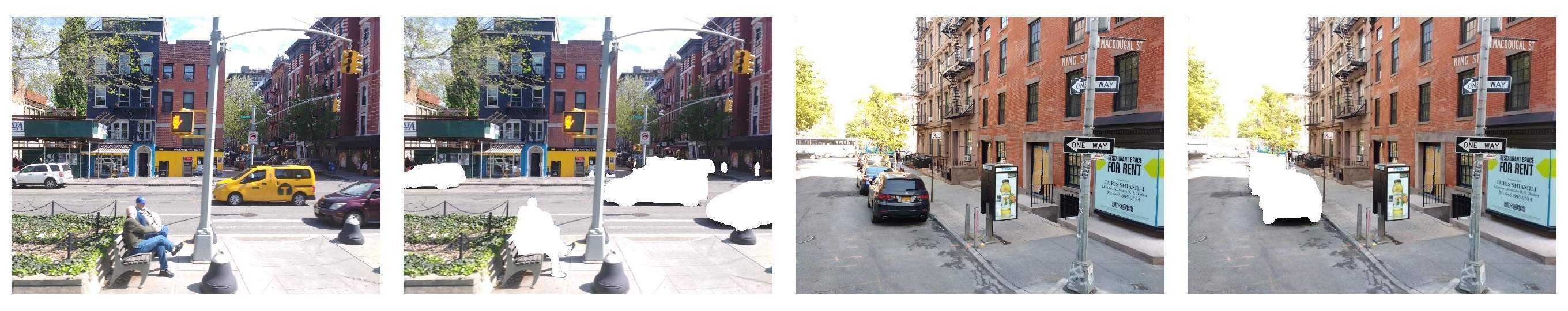}
    \vspace{-2mm}
    \caption{Raw images vs. Anonymized images. \label{fig:raw_anony}}
    \vspace{-5mm}
\end{figure}

Our dataset is named \ourname.  It contains images recorded in Manhattan, New York from May 2016 to March 2017. The images were recorded with GPS tags by smartphone cameras installed on the front, back, and side parts of (undisclosed) fleet cars with auto-exposure\footnote{The raw data was sampled from a larger dataset provided by Carmera.}. The dataset contains both side-view images and front-view images. There are 100,500 side-view and 101,290 front-view raw images, each with a $640\times 480$ resolution. On the basis of raw images, we use MSeg \cite{MSeg_2020_CVPR}, a semantic segmentation method, to replace moving objects such as people and cars with white pixels. Fig. \ref{fig:raw_anony} compares anonymized and raw images.

\begin{figure*}[t]
    \centering
    \subfigure[During vs. After construction.]{
        \includegraphics[width=0.3\linewidth]{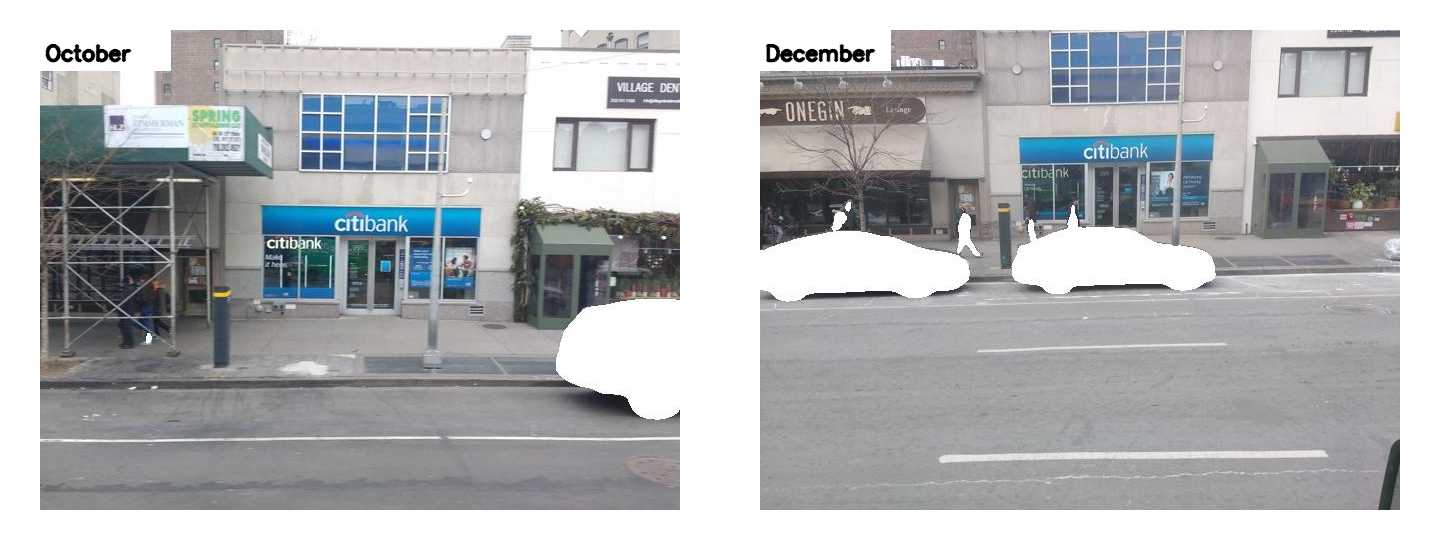}
        \label{fig:construction}
    }
    \subfigure[Summer vs. Winter.]{
        \includegraphics[width=0.3\linewidth]{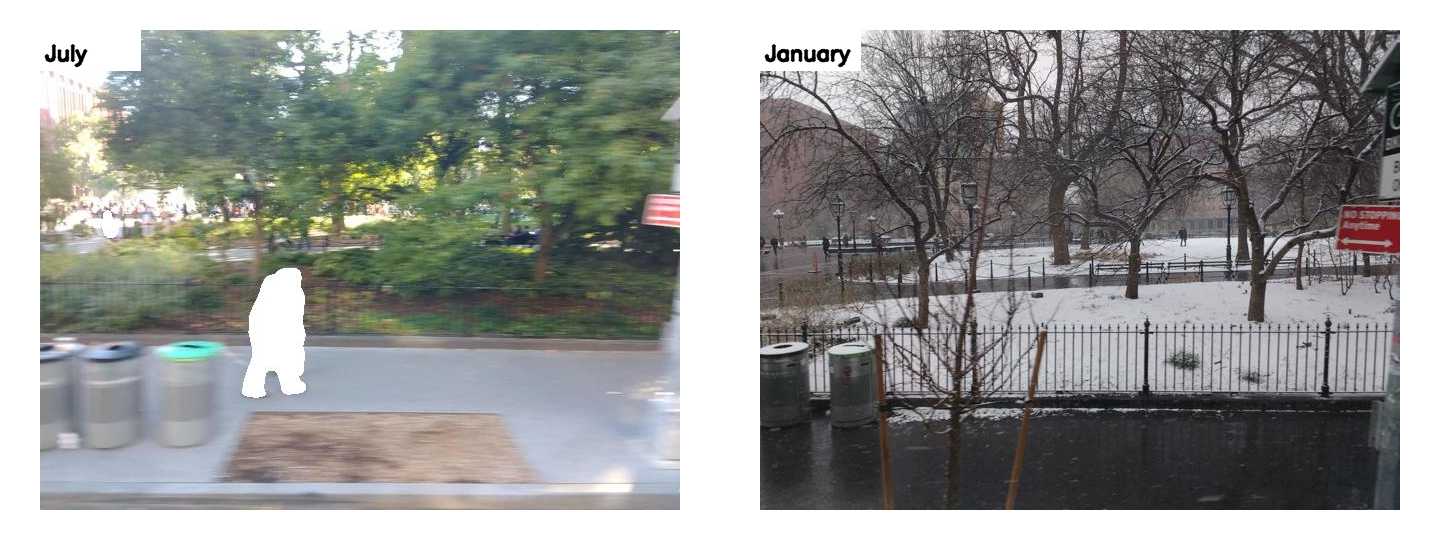}
        \label{fig:park}
    }
    \subfigure[Without vs. With motion blur.]{
        \includegraphics[width=0.3\linewidth]{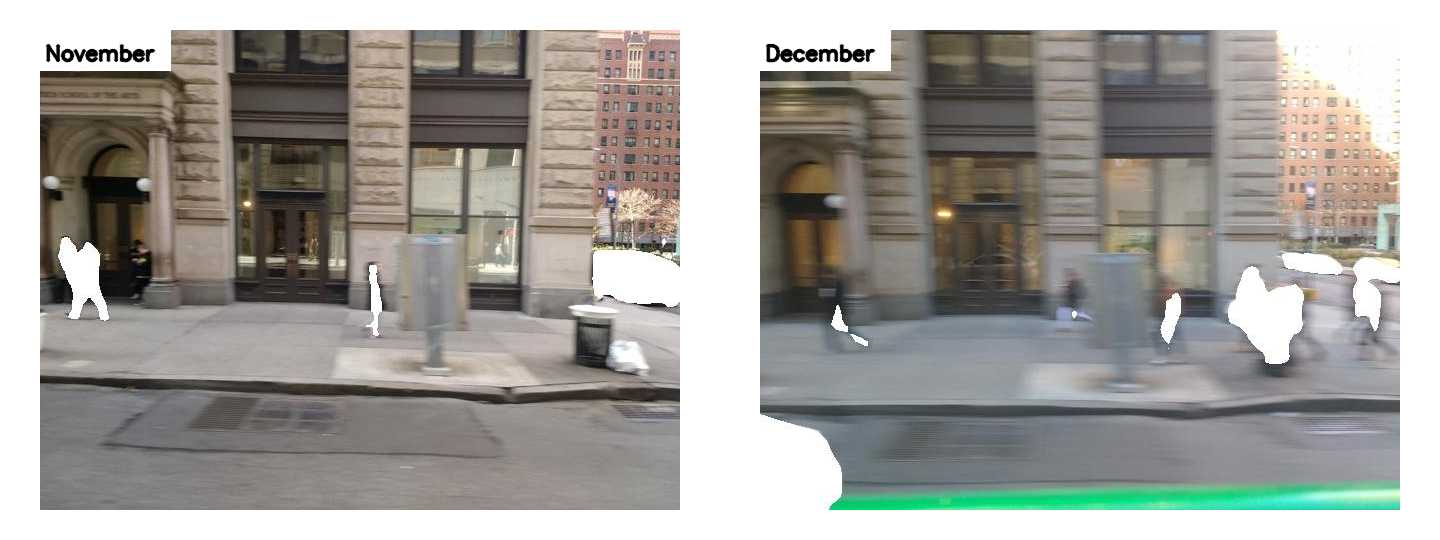}
        \label{fig:blur}
    }
    \subfigure[Image GPS locations.]{
        \includegraphics[width=0.22\linewidth]{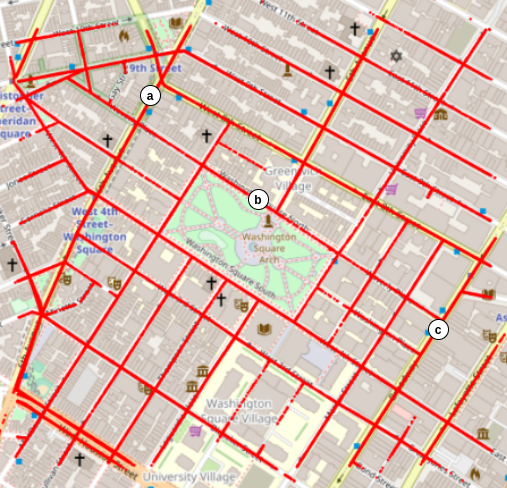}
        \label{fig:trajectories}
    }
    \subfigure[Frequency of front-view images.]{
        \includegraphics[width=0.22\linewidth]{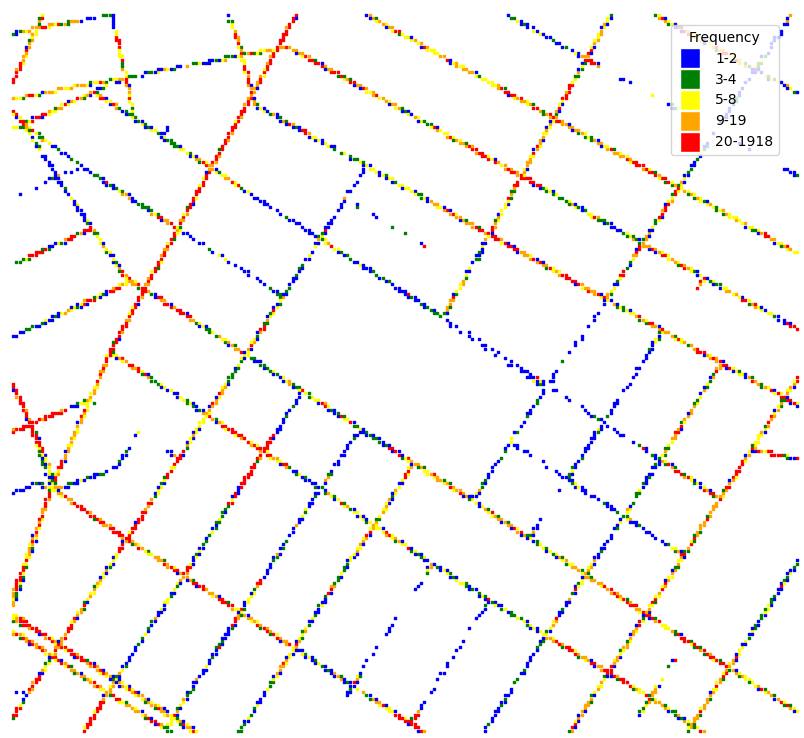}
        \label{fig:Frequency_of_front_view}
    }
    \subfigure[Frequency of side-view images.]{
        \includegraphics[width=0.22\linewidth]{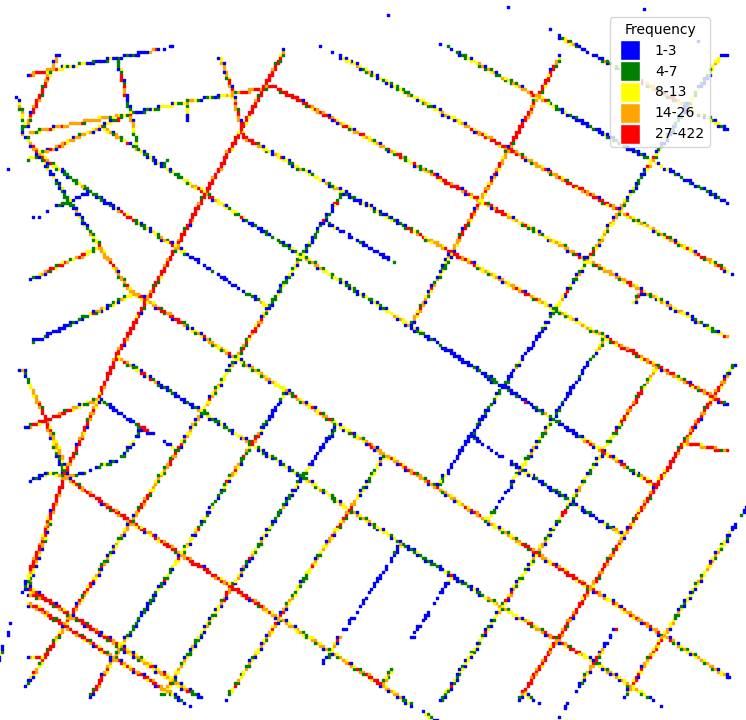}
        \label{fig:Frequency_of_side_view}
    }
    \caption{Dataset visualization of \ourname~with respect to the image capturing location. The locations of (a)-(c) are highlighted in (d).}
    \vspace{-5mm}
\end{figure*}

The images were recorded on streets around Washington Square Park. The trajectories of the locations where the images were recorded are shown in Fig. \ref{fig:trajectories}. Since the cameras were placed on fleet cars, and their routes were random, the frequencies of locations where the images were taken are different. The frequencies of the locations where the side-view and front-view images were recorded are shown in Fig. \ref{fig:Frequency_of_side_view} and Fig. \ref{fig:Frequency_of_front_view} respectively.

Fig. \ref{fig:time_figure} shows the time distribution. Since it contains images captured from May 2016 to March 2017, our dataset includes all four seasons. Therefore, it contains various changes of weather, illumination, vegetation, and road construction. As shown in Fig. \ref{fig:same_location}, we can see image changes at the same location as the season changes.

\textbf{Difficulty Level: }We assign each side-view query image a difficulty level of easy, medium, or hard. First, we extract SIFT~\cite{790410} features for each image. Then for each query image, we find the top-8 closest side-view training images by GPS coordinates. The query image and its top-8 closest images form eight image pairs. We use RANSAC to compute a fundamental matrix and the number of inliers for each pair of images. We use three intervals to measure the difficulty level of matching each pair based on each pair's number of inliers points: 0-19 (hard), 20-80 (medium), $>$80 (easy). The interval values are determined by artificially viewing the image pairs and checking the similarity of the image pairs. The difficulty level of each side-view query image is the most common difficulty level of its eight pairs.

\textbf{Uniqueness: } Our dataset is unique in two ways. First, comparing to front-view images where sky and road surfaces occupy large areas, side-view images focus on street views such as shop signs and metro entrances. 
Second, we include image
anonymization to protect the privacy of pedestrians and cars. In the meantime, anonymized images provide VPR algorithms static and environment-only information, getting rid of moving objects and pedestrians.

\begin{figure}
    \centering
    \subfigure[Sequential view overlaps.]{
        \includegraphics[width=0.45\columnwidth]{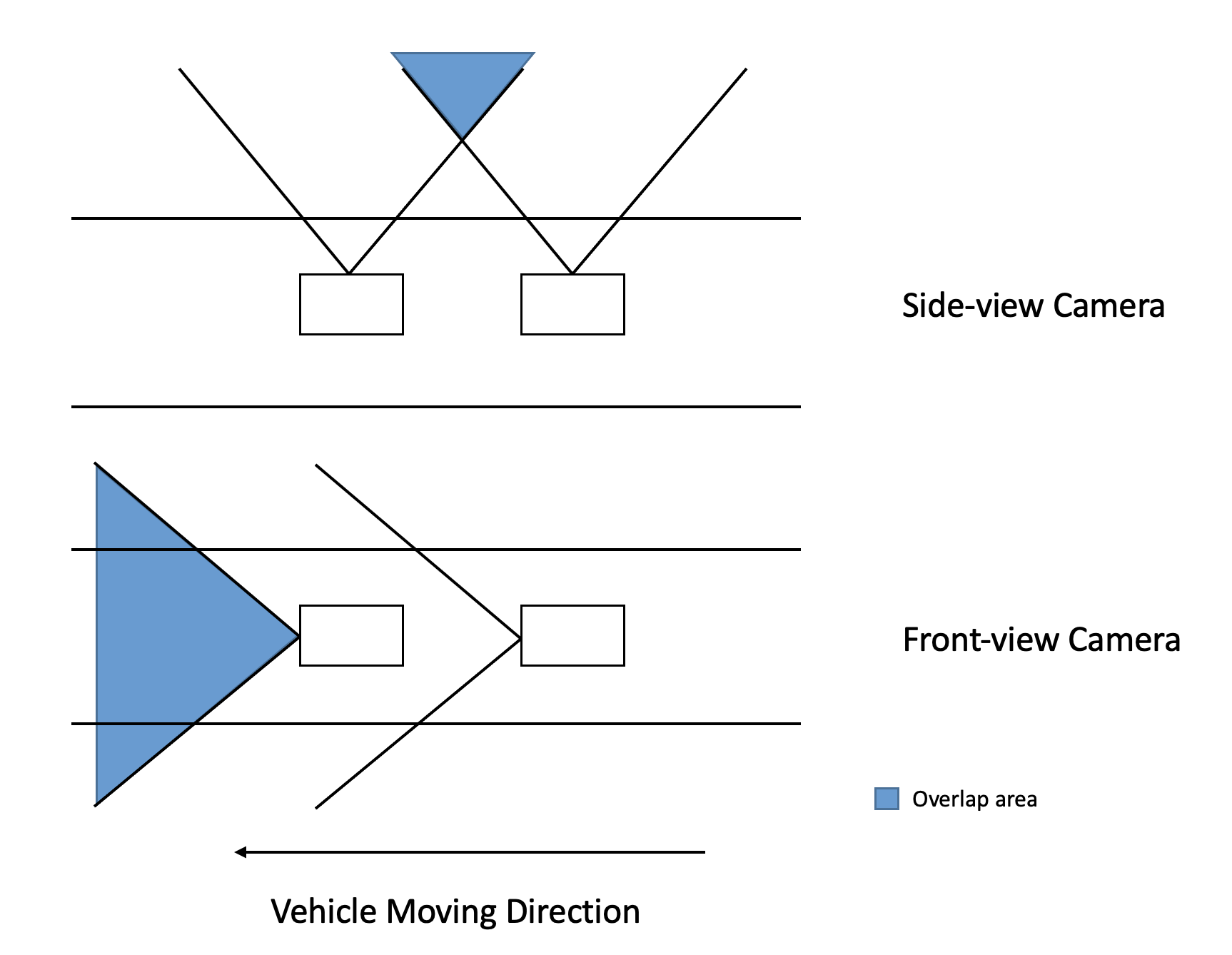}
        \label{fig:explanation}
        \vspace{-5mm}
    }
    \subfigure[Percentage of blurry images.]{
        \includegraphics[width=0.45\columnwidth]{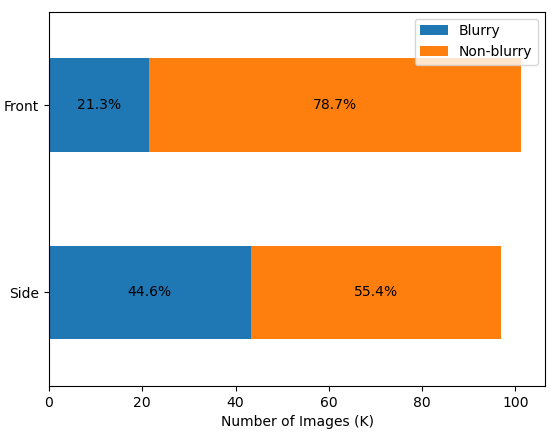}
        \label{fig:blurry_percentage}
    }
    \caption{Further explanation and illustration of our dataset.}
    \label{fig:combination}
    \vspace{-5mm}
\end{figure}

\textbf{Front-view vs. Side-view}:
\textit{We hypothesize that side-view images are more challenging than front-view images for existing VPR methods, based on several theoretical reasons and observations}. First, as illustrated in Fig.~\ref{fig:explanation}, since the images were taken on a moving vehicle at a low frequency, the overlap of two sequential side-view images would only occupy a limited part of the whole image. In contrast, the overlap of two sequential front-view images occupies more space. Furthermore, if the road is narrow and stores are very close to the vehicle, the spatial area covered in the side-view image is less than that covered in the front-view image at the same location.
Besides, trees may impact the success rate for side-view images because trees will block features such as banners and entrances. Most trees look the same and thus may cause the uncertainty of localization.

Another important reason is that images are taken on a moving vehicle, suffering from motion blur. When the spatial area coverage of side-view images is less than front-view images, the motion blur effect is more serious on side-view images. We set a threshold for the variance of Laplacian to detect the blurriness of an image\footnote{\url{https://www.pyimagesearch.com/2015/09/07/blur-detection-with-opencv/}}. The result in Fig.~\ref{fig:blurry_percentage} right
shows that side-view images have a larger percentage of blurry images than front-view images. We think blurry images are more challenging for VPR and we will analyze the relationship further in the experiment section.

\textbf{Other Challenges: }In addition to the side-view images, there are several other challenges in our dataset. Because our dataset is one-year long, the images taken at the same location have artificial or natural differences. First, Fig.~\ref{fig:construction} left was taken in October 2016 with sideway constructions and the right was taken in December 2016 after the construction. At some locations, the construction may cover the whole image. Second, different seasons cause different appearances at the same location. Fig. \ref{fig:park} left was taken in summer, July 2016, and the right was taken in winter, January 2017. In this case, the vegetation in Washington Square Park had changed a lot and snow was covering the ground in winter. Furthermore, if the vehicle was moving fast, the images taken by the vehicle will be blurry (Fig. \ref{fig:blur}). Although two images were taken at the same location, the blurry one will cause more difficulty during VPR.

\section{Benchmark Experiments}
\subsection{Settings}

We selected five classical as well as state-of-the-art descriptors and methods for evaluation of visual place recognition performance on the {\ourname} dataset. 

\textbf{Dataset: }We randomly split both front-view and side-view images into training, validation and testing sections by 80\%, 5\%, and 15\% respectively. For each view direction, both anonymized and raw images share the same split result. All images are resized to $640\times 480$. We also use Python module utm to convert GPS coordinates to UTM coordinates for more precise distance calculation between two locations.

\textbf{VLAD+SURF: }We use Vector of Locally Aggregated Descriptors(VLAD)~\cite{jegou2010aggregating} to aggregate speeded up robust features (SURF)~\cite{bay2006surf} descriptors for image retrieval. Through experiments, we find the optimal cluster number is 32 within 8, 16, 32, and 64, by using MiniBatchKmeans with batch size at 5000. This cluster number gives high accuracy and acceptable training time. The training of 77608 images took about 8 hours on CPU with 64 GB available memory.

\textbf{VLAD+SuperPoint: }We use SuperPoint model pre-trained on MS-COCO generic image dataset~\cite{detone18superpoint}. We use nVidia RTX 2080S to extract SuperPoint features. Then we use VLAD to aggregate SuperPoint descriptors for image retrieval. We set the cluster number at 32, just as we do in VLAD+SURF, by using MiniBatchKmeans with batch size at 100. Notice the dimension of SuperPoint descriptors is larger than the dimension of SURF descriptors. The training of 77608 images took about 20 hours on CPU and GPU with 64 GB available memory.

\textbf{NetVLAD: }We directly use the pre-trained model weight for 30 epochs on Pittsburgh-250k datasets~\cite{Torii-CVPR2013} to complete our testing. For the hardware, the CPU we adapt is Intel\textregistered\, Core\textsuperscript{TM} i7-8700k, and the GPU we use is NVIDIA GEFORCE GTX 1080 TI. We first complete an initial clustering on training data to find out the centroids used for the testing process. The input testing data with the extracted deep feature are assigned to different clusters afterward. The batch size during testing is 24.

\textbf{PoseNet:} We use PoseNet model with ResNet34 as the base architecture~\cite{kendall2015posenet}. For training, PoseNet requires the Cartesian coordinates of images as input besides images themselves. So we gather latitude and longitude information of training images from the camera. We convert latitude and longitude to universal transverse mercator (UTM) coordinates to improve the accuracy of PoseNet’s estimation of images’ relative position. We use images with normalized UTM coordinates as the input to PoseNet. The GPU we use is NVIDIA GEFORCE RTX 2080S. The batch size during training is 32. For testing, PoseNet outputs the estimation of normalized UTM coordinates of the query images, which are used for evaluation.

\textbf{DBoW:} We use Distributed Bag of Words (DBoW) model\footnote{\url{https://github.com/rmsalinas/DBow3}}. We choose Oriented FAST and Rotated BRIEF (ORB) \cite{orb2011} descriptors for representing features. We use DBoW to generate a vocabulary constructed by ORB descriptors of training and test images. For testing, We generate the top-5 retrieval images by using DBoW3 to generate a score between each training and test images and selecting the top-5 scores for each test image. We run the testing process with multi-thread for efficiency.

\textbf{Evaluation: } Following~\cite{yu2018vlase}, we measure both top-1 and top-5 retrieval accuracy under four distance thresholds (5, 10, 15, 20 meters). If any of the top-k retrieval images are within the range of the distance from the query image, we count it as a successful retrieval. The top-k ranking is based on the similarity between image features calculated by VPR algorithms. In essence, this evaluation metric is similar to the more common precision-recall curve.

\subsection{Results}

\begin{figure}[ht]
    \centering
    \subfigure[Top-5 retrieval results.]{
        \includegraphics[width=8.5cm]{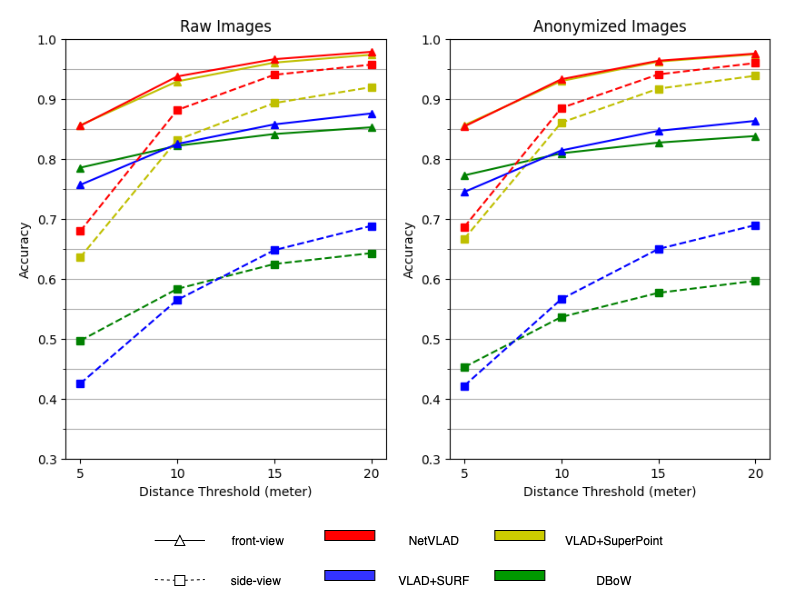}
        \label{fig:results}
    }
    \subfigure[Top-5 retrieval results in terms of difficulty level.]{
        \includegraphics[width=8.5cm]{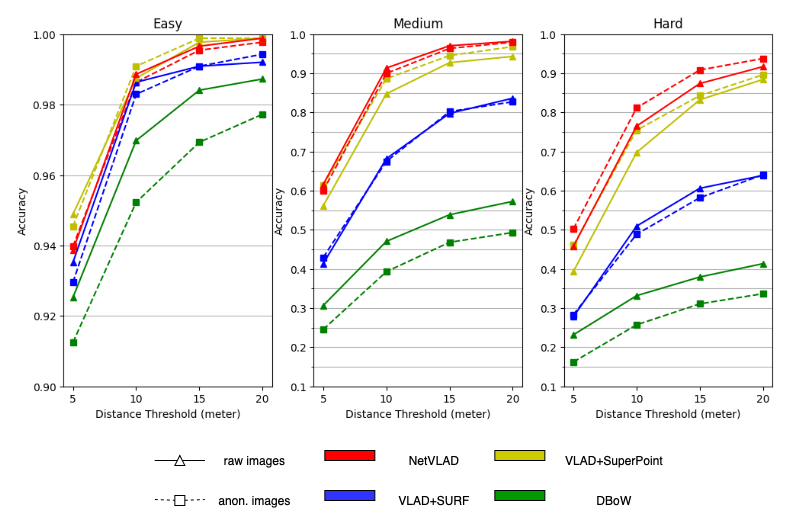}
        \label{fig:level_results}
    }
    \caption{Main results of the benchmark.}\label{fig:main-results}
    \vspace{-5mm}
\end{figure}

Figure \ref{fig:results} shows our main results focused on the performance of non-/anonymized front-/side-view datasets, showing 
the influence of anonymization and view directions.

\begin{figure*}[t]
    \centering
    \includegraphics[width=0.7\linewidth]{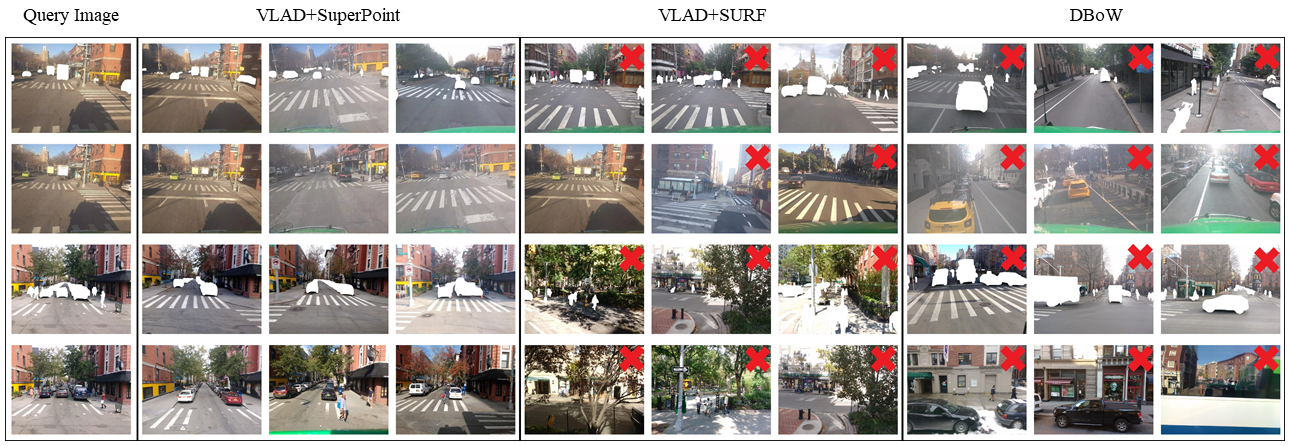}

    \vspace{-3mm}
    \caption{We visualize VPR results only for VLAD+SuperPoint, VLAD+SURF, and DBOW, because the result of NetVLAD is very similar to VLAD+SuperPoint. We randomly picked one location (more locations results in the supplementary). The first row is for the anonymized front-view query. The second row is for the corresponding raw front-view query. The third row is for the side-view query taken at the same location as the front-view query. The last row is for the corresponding raw side-view query. The red cross means the location of the retrieval image is not in the distance threshold (10 meters). We show the top 3 retrievals for each method.}
    \label{fig:visual_result}
    \vspace{-3mm}
\end{figure*}

\textbf{Performance: }Fig. \ref{fig:main} and Fig. \ref{fig:results} shows the result of our experiments. Obviously, the result of the top 5 retrieval image accuracy is higher than the top 1 retrieval image accuracy, with an average 10\% higher. And when using VLAD to aggregate descriptors, the result of SuperPoint descriptors is much more accurate than the result of SURF descriptors. In both two figures, the accuracy of NetVLAD is the highest, followed by VLAD+SuperPoint and VLAD+SURF. The last is the DBoW method. We may attribute the low accuracy of DBoW to the unsustainability of ORB features. PoseNet outputs a GPS coordinate and using that coordinate we can find the closest top 1 retrieval image, and through experiments, the accuracy of PoseNet is 15.3\% and 37.5\% when the distance threshold is 5 and 10 meters respectively. \textit{Due to the low performance of PoseNet, we omit it in other experiments and do not plot the results}. We also calculate the accuracy in terms of difficulty level as mentioned before. Fig. \ref{fig:level_results} shows the result of top 5 retrieval in different difficulty level. The visual result is shown in Fig \ref{fig:visual_result}. Clearly, we can see VLAD+SuperPoint has better performance than VLAD+SURF and DBoW.

\textbf{Anonymization: }From Fig. \ref{fig:main}, we can draw the conclusion that the anonymization does not have a large impact on the visual place recognition result, either of front-view dataset or side-view dataset. The anonymization, however, has little influence on the accuracy of some methods. For example, VLAD+SuperPoint gets 1.1\% increase on average, while DBoW and VLAD+SURF have around 2.1\% and  3.4\% decrease on average respectively. Therefore, when doing experiments on visual place recognition, we can anonymize raw images to protect the privacy of people and cars.

\textbf{View Direction: }View direction does have a conspicuous influence on retrieval accuracy. In Fig.~\ref{fig:main}, the accuracies calculated from front-view images are higher than those from side-view images. This phenomenon happens in every method, indicating that front-view images is less challenging than side-view images in VPR, as we expected previously. 

Next, we provide some in-depth analysis according to the reasons we hypothesized in Fig. ~\ref{fig:explanation}. We first define that \textit{a query image is successfully retrieved if any of the top 5 retrieval images is within 5 meters of the query image}.

\begin{figure}
    \centering
    \includegraphics[width=0.7\columnwidth]{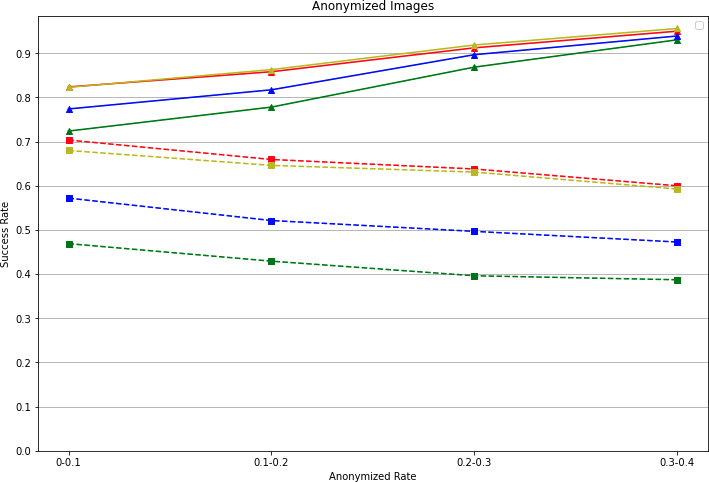}
    \vspace{-3mm}
    \caption{VPR success rate vs. anonymized rate. The same legend as Fig. \ref{fig:main-results}.}
    \label{fig:success_rate_by_anonymization}
    \vspace{-3mm}
\end{figure}

\textbf{Dynamic Objects:} Fig. \ref{fig:success_rate_by_anonymization} shows the success rate of query images vs. different anonymization rates. We calculate the anonymization rate using the percentage of white pixels in the image. As the anonymization rate increases, the success rate for the front view increases while the success rate for the side view decreases. We propose a hypothesis for why this happens: in front view, dynamic objects such as cars and pedestrians are noisy signals for VPR. After anonymizing dynamic objects, VPR algorithms focus on the street features, which increases the success rate. However, in side view, those dynamic objects may block many street features. A higher anonymization rate means more street features blocked, which decreases the success rate.

\begin{figure}
    \centering
    \includegraphics[width=0.7\columnwidth]{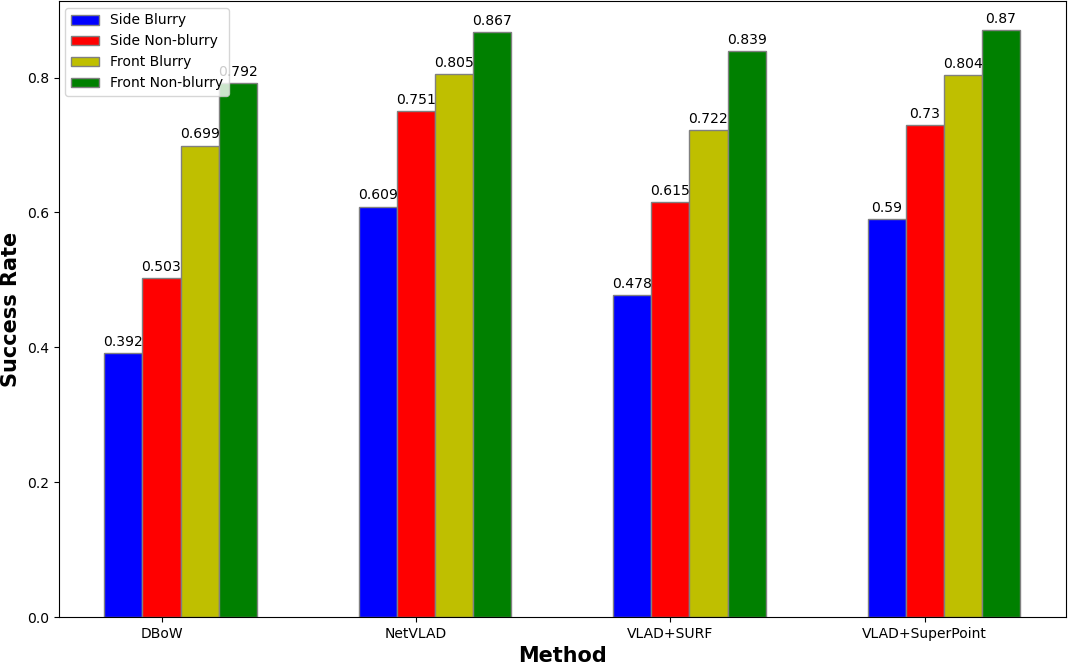}
    \vspace{-3mm}
    \caption{VPR success rate vs. image blurriness.}
    \label{fig:success_rate_by_blurry}
    \vspace{-5mm}
\end{figure}

\textbf{Motion Blur:} Fig. \ref{fig:success_rate_by_blurry} shows the relationship among success rate, VPR method, view directions, and blurriness. It clearly shows that side-view images are more challenging than front-view images, and blurry images are more challenging than non-blurry ones. Moreover, given the same blurriness condition, side-view is still more challenging than front-view. We believe this is also due to the reason we hypothesized before: front-view contains more view overlaps among neighboring images, thus better VPR feature matching.

\begin{figure}
    \centering
    \includegraphics[width=0.65\columnwidth]{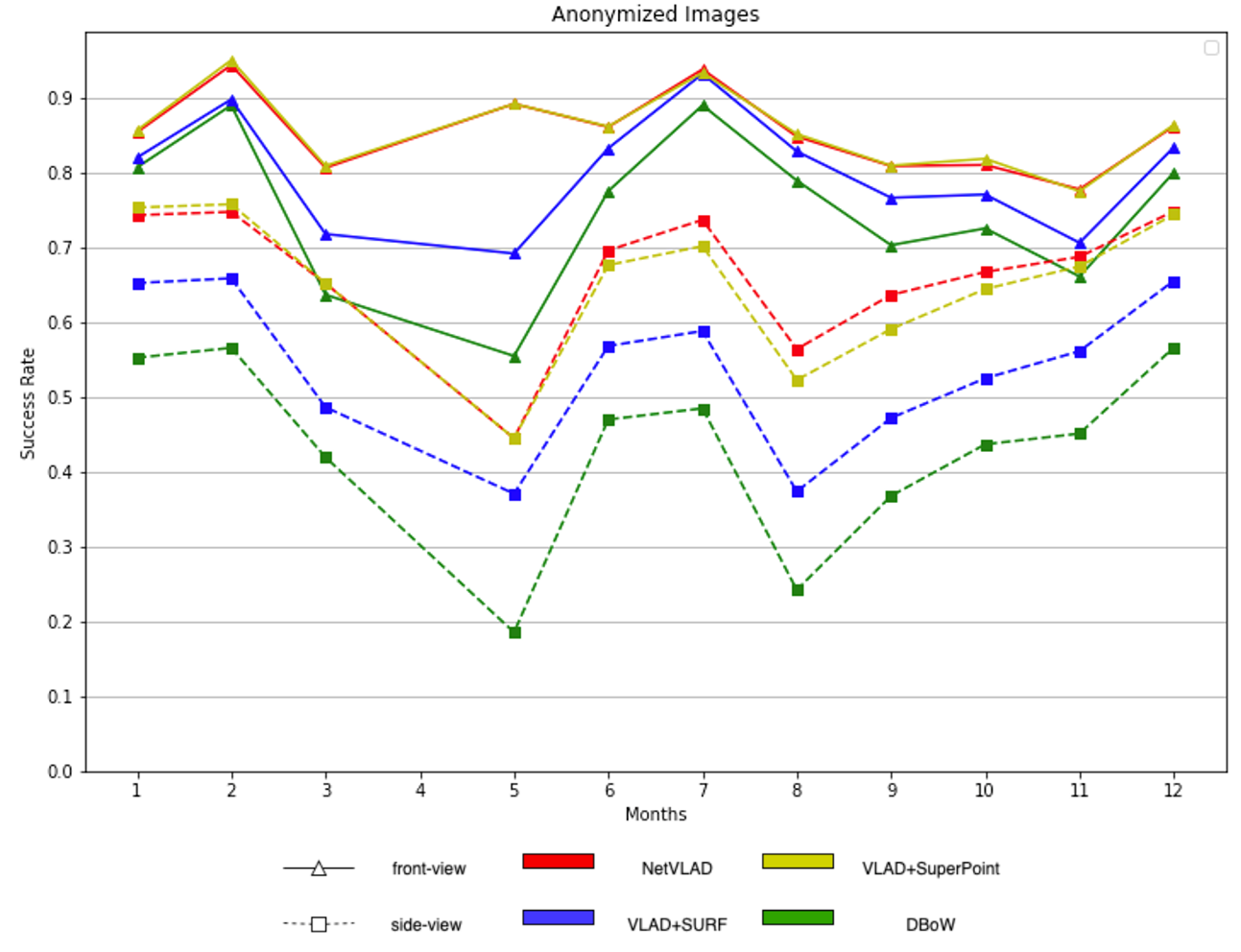}
    \vspace{-3mm}
    \caption{VPR success rate vs. query image month.}
    \label{fig:success_rate_by_month}
    \vspace{-5mm}
\end{figure}

\textbf{Seasonal Change: } In addition, we analyzed the success rate of query images in different months as shown in Fig.~\ref{fig:success_rate_by_month}. This result further confirmed in each month our observation that the side view is more challenging than the front view. The fluctuations between months also reflect some season/weather related influences on VPR performance.

\section{Conclusions}

After the large-scale experiments and analysis, we can finally answer our questions with confidence. Are side-view images more challenging for VPR than front-view ones? \textit{Yes, and the performance drops of all VPR methods are significant}, although the dataset has no significant spatial/temporal differences on the distribution of images captured from the two view directions. Would our data anonymization significantly affect the performance of existing VPR algorithms? \textit{No, and for some methods, the anonymization could even bring marginal improvements}, potentially due to the removal of those VPR noises. Our future work includes benchmarking more VPR methods and with geometric verification.

\section*{Acknowledgments}
We would like to thank Carmera for providing the raw NYC image data set that we used for creating the NYU-VPR. This research is funded by the Connected Cities for Smart Mobility towards Accessible and Resilient Transportation (C2SMART), a Tier 1 University Center awarded by U.S. Department of Transportation under contract 69A3351747124.

\addtolength{\textheight}{-0.2cm}   %

\bibliographystyle{IEEEtranN}
\bibliography{ai4ce-tpl}

% Generated by IEEEtranN.bst, version: 1.14 (2015/08/26)
\begin{thebibliography}{26}
\providecommand{\natexlab}[1]{#1}
\providecommand{\url}[1]{#1}
\csname url@samestyle\endcsname
\providecommand{\newblock}{\relax}
\providecommand{\bibinfo}[2]{#2}
\providecommand{\BIBentrySTDinterwordspacing}{\spaceskip=0pt\relax}
\providecommand{\BIBentryALTinterwordstretchfactor}{4}
\providecommand{\BIBentryALTinterwordspacing}{\spaceskip=\fontdimen2\font plus
\BIBentryALTinterwordstretchfactor\fontdimen3\font minus
  \fontdimen4\font\relax}
\providecommand{\BIBforeignlanguage}[2]{{%
\expandafter\ifx\csname l@#1\endcsname\relax
\typeout{** WARNING: IEEEtranN.bst: No hyphenation pattern has been}%
\typeout{** loaded for the language `#1'. Using the pattern for}%
\typeout{** the default language instead.}%
\else
\language=\csname l@#1\endcsname
\fi
#2}}
\providecommand{\BIBdecl}{\relax}
\BIBdecl

\bibitem[Mirowski et~al.(2019)Mirowski, Banki-Horvath, Anderson, Teplyashin,
  Hermann, Malinowski, Grimes, Simonyan, Kavukcuoglu, Zisserman,
  et~al.]{mirowski2019streetlearn}
P.~Mirowski, A.~Banki-Horvath, K.~Anderson, D.~Teplyashin, K.~M. Hermann,
  M.~Malinowski, M.~K. Grimes, K.~Simonyan, K.~Kavukcuoglu, A.~Zisserman
  \emph{et~al.}, ``The streetlearn environment and dataset,'' \emph{arXiv
  preprint arXiv:1903.01292}, 2019.

\bibitem[Zamir and Shah(2014)]{zamir2014image}
A.~R. Zamir and M.~Shah, ``Image geo-localization based on multiplenearest
  neighbor feature matching usinggeneralized graphs,'' \emph{{IEEE} Trans.
  Pattern Anal. Mach. Intell.}, vol.~36, no.~8, pp. 1546--1558, 2014.

\bibitem[S{\"u}nderhauf et~al.(2013)S{\"u}nderhauf, Neubert, and
  Protzel]{sunderhauf2013we}
N.~S{\"u}nderhauf, P.~Neubert, and P.~Protzel, ``Are we there yet? challenging
  seqslam on a 3000 km journey across all four seasons,'' in \emph{Proc. IEEE
  Int'l Conf. Robotics and Automation (ICRA)}, 2013, p. 2013.

\bibitem[Sue()]{Suenderhauf_vprice_2015}
\BIBentryALTinterwordspacing
``The {VPRiCE} {Challenge} 2015 – {Visual} {Place} {Recognition} in
  {Changing} {Environments} - {Public} - {Confluence}.'' [Online]. Available:
  \url{https://roboticvision.atlassian.net/wiki/spaces/PUB/pages/14188617/The+VPRiCE+Challenge+2015+Visual+Place+Recognition+in+Changing+Environments}
\BIBentrySTDinterwordspacing

\bibitem[Torii et~al.(2015)Torii, Arandjelovic, Sivic, Okutomi, and
  Pajdla]{torii201524}
A.~Torii, R.~Arandjelovic, J.~Sivic, M.~Okutomi, and T.~Pajdla, ``24/7 place
  recognition by view synthesis,'' in \emph{Proc. IEEE Conf. Computer Vision
  and Pattern Recognition (CVPR)}, 2015, pp. 1808--1817.

\bibitem[Torii et~al.(2013)Torii, Sivic, Pajdla, and Okutomi]{Torii-CVPR2013}
A.~Torii, J.~Sivic, T.~Pajdla, and M.~Okutomi, ``Visual place recognition with
  repetitive structures,'' in \emph{Proc. IEEE Conf. Computer Vision and
  Pattern Recognition (CVPR)}, 2013.

\bibitem[Geiger et~al.(2013)Geiger, Lenz, Stiller, and Urtasun]{Geiger2013IJRR}
A.~Geiger, P.~Lenz, C.~Stiller, and R.~Urtasun, ``Vision meets robotics: The
  kitti dataset,'' \emph{Int'l J. Robotics Research}, 2013.

\bibitem[Choi et~al.(2015)Choi, Kim, Park, Hwang, Yoon, In, and Kweon]{parkall}
Y.~Choi, N.~Kim, K.~Park, S.~Hwang, J.~S. Yoon, Y.~In, and I.~Kweon, ``All-day
  visual place recognition: Benchmark dataset and baseline,'' in \emph{Workshop
  on Visual Place Recognition in Changing Environments}, 06 2015.

\bibitem[Maddern et~al.(2017)Maddern, Pascoe, Linegar, and
  Newman]{maddern20171}
W.~Maddern, G.~Pascoe, C.~Linegar, and P.~Newman, ``1 year, 1000 km: The oxford
  robotcar dataset,'' \emph{Int'l J. Robotics Research}, vol.~36, no.~1, pp.
  3--15, 2017.

\bibitem[Warburg et~al.(2020)Warburg, Hauberg, Lopez-Antequera, Gargallo,
  Kuang, and Civera]{warburg2020mapillary}
F.~Warburg, S.~Hauberg, M.~Lopez-Antequera, P.~Gargallo, Y.~Kuang, and
  J.~Civera, ``Mapillary street-level sequences: A dataset for lifelong place
  recognition,'' in \emph{Proc. IEEE Conf. Computer Vision and Pattern
  Recognition (CVPR)}, 2020, pp. 2626--2635.

\bibitem[Carlevaris-Bianco et~al.(2016)Carlevaris-Bianco, Ushani, and
  Eustice]{carlevaris2016university}
N.~Carlevaris-Bianco, A.~K. Ushani, and R.~M. Eustice, ``University of michigan
  north campus long-term vision and lidar dataset,'' \emph{The International
  Journal of Robotics Research}, vol.~35, no.~9, pp. 1023--1035, 2016.

\bibitem[Speciale et~al.(2019)Speciale, Schonberger, Kang, Sinha, and
  Pollefeys]{speciale2019privacy}
P.~Speciale, J.~L. Schonberger, S.~B. Kang, S.~N. Sinha, and M.~Pollefeys,
  ``Privacy preserving image-based localization,'' in \emph{Proc. IEEE Conf.
  Computer Vision and Pattern Recognition (CVPR)}, 2019, pp. 5493--5503.

\bibitem[Lambert et~al.(2020)Lambert, Liu, Sener, Hays, and
  Koltun]{MSeg_2020_CVPR}
J.~Lambert, Z.~Liu, O.~Sener, J.~Hays, and V.~Koltun, ``{MSeg}: A composite
  dataset for multi-domain semantic segmentation,'' in \emph{Proc. IEEE Conf.
  Computer Vision and Pattern Recognition (CVPR)}, 2020.

\bibitem[Arandjelovic et~al.(2016)Arandjelovic, Gronat, Torii, Pajdla, and
  Sivic]{arandjelovic2016netvlad}
R.~Arandjelovic, P.~Gronat, A.~Torii, T.~Pajdla, and J.~Sivic, ``Netvlad: Cnn
  architecture for weakly supervised place recognition,'' in \emph{Proc. IEEE
  Conf. Computer Vision and Pattern Recognition (CVPR)}, 2016, pp. 5297--5307.

\bibitem[Kendall et~al.(2015)Kendall, Grimes, and Cipolla]{kendall2015posenet}
A.~Kendall, M.~Grimes, and R.~Cipolla, ``Posenet: A convolutional network for
  real-time 6-dof camera relocalization,'' in \emph{Proc. IEEE Int'l Conf.
  Computer Vision (ICCV)}, 2015, pp. 2938--2946.

\bibitem[{Chanc\'an} et~al.(2020){Chanc\'an}, {Hernandez-Nunez}, {Narendra},
  {Barron}, and {Milford}]{chancan2020hybrid}
M.~{Chanc\'an}, L.~{Hernandez-Nunez}, A.~{Narendra}, A.~B. {Barron}, and
  M.~{Milford}, ``A hybrid compact neural architecture for visual place
  recognition,'' \emph{IEEE Robotics and Automation Letters}, vol.~5, no.~2,
  pp. 993--1000, April 2020.

\bibitem[Chen et~al.(2017)Chen, Jacobson, S{\"u}nderhauf, Upcroft, Liu, Shen,
  Reid, and Milford]{chen2017deep}
Z.~Chen, A.~Jacobson, N.~S{\"u}nderhauf, B.~Upcroft, L.~Liu, C.~Shen, I.~Reid,
  and M.~Milford, ``Deep learning features at scale for visual place
  recognition,'' in \emph{Proc. IEEE Int'l Conf. Robotics and Automation
  (ICRA)}.\hskip 1em plus 0.5em minus 0.4em\relax IEEE, 2017, pp. 3223--3230.

\bibitem[G\'alvez-L\'opez and Tard\'os(2012)]{GalvezTRO12}
D.~G\'alvez-L\'opez and J.~D. Tard\'os, ``Bags of binary words for fast place
  recognition in image sequences,'' \emph{{IEEE} Trans. Robotics}, vol.~28,
  no.~5, pp. 1188--1197, October 2012.

\bibitem[J{\'e}gou et~al.(2010)J{\'e}gou, Douze, Schmid, and
  P{\'e}rez]{jegou2010aggregating}
H.~J{\'e}gou, M.~Douze, C.~Schmid, and P.~P{\'e}rez, ``Aggregating local
  descriptors into a compact image representation,'' in \emph{Proc. IEEE Conf.
  Computer Vision and Pattern Recognition (CVPR)}.\hskip 1em plus 0.5em minus
  0.4em\relax IEEE, 2010, pp. 3304--3311.

\bibitem[Sattler et~al.(2012)Sattler, Leibe, and Kobbelt]{sattler2012improving}
T.~Sattler, B.~Leibe, and L.~Kobbelt, ``Improving image-based localization by
  active correspondence search,'' in \emph{Proc. European Conf. Computer Vision
  (ECCV)}.\hskip 1em plus 0.5em minus 0.4em\relax Springer, 2012, pp. 752--765.

\bibitem[{Rublee} et~al.(2011){Rublee}, {Rabaud}, {Konolige}, and
  {Bradski}]{orb2011}
E.~{Rublee}, V.~{Rabaud}, K.~{Konolige}, and G.~{Bradski}, ``Orb: An efficient
  alternative to sift or surf,'' in \emph{Proc. IEEE Int'l Conf. Computer
  Vision (ICCV)}, 2011, pp. 2564--2571.

\bibitem[{Mur-Artal} et~al.(2015){Mur-Artal}, {Montiel}, and
  {Tardós}]{7219438}
R.~{Mur-Artal}, J.~M.~M. {Montiel}, and J.~D. {Tardós}, ``Orb-slam: A
  versatile and accurate monocular slam system,'' \emph{{IEEE} Trans.
  Robotics}, vol.~31, no.~5, pp. 1147--1163, 2015.

\bibitem[Bay et~al.(2006)Bay, Tuytelaars, and Van~Gool]{bay2006surf}
H.~Bay, T.~Tuytelaars, and L.~Van~Gool, ``Surf: Speeded up robust features,''
  in \emph{Proc. European Conf. Computer Vision (ECCV)}.\hskip 1em plus 0.5em
  minus 0.4em\relax Springer, 2006, pp. 404--417.

\bibitem[Yu et~al.(2018)Yu, Chaturvedi, Feng, Taguchi, Lee, Fernandes, and
  Ramalingam]{yu2018vlase}
X.~Yu, S.~Chaturvedi, C.~Feng, Y.~Taguchi, T.-Y. Lee, C.~Fernandes, and
  S.~Ramalingam, ``{VLASE}: Vehicle localization by aggregating semantic
  edges,'' \emph{Proc. IEEE/RSJ Int'l Conf. Intelligent Robots and Systems
  (IROS)}, 2018.

\bibitem[DeTone et~al.(2018)DeTone, Malisiewicz, and
  Rabinovich]{detone18superpoint}
\BIBentryALTinterwordspacing
D.~DeTone, T.~Malisiewicz, and A.~Rabinovich, ``Superpoint: Self-supervised
  interest point detection and description,'' in \emph{CVPR Deep Learning for
  Visual SLAM Workshop}, 2018. [Online]. Available:
  \url{http://arxiv.org/abs/1712.07629}
\BIBentrySTDinterwordspacing

\bibitem[{Lowe}(1999)]{790410}
D.~G. {Lowe}, ``Object recognition from local scale-invariant features,'' in
  \emph{Proc. IEEE Int'l Conf. Computer Vision (ICCV)}, vol.~2, 1999, pp.
  1150--1157 vol.2.

\end{thebibliography}

\end{document}